\def\year{2021}\relax
\documentclass[letterpaper]{article} 
\usepackage{aaai21}  
\usepackage{times}  
\usepackage{helvet} 
\usepackage{courier}  
\usepackage[hyphens]{url}  
\usepackage{graphicx} 
\usepackage{amsmath}
\usepackage{amsfonts}
\usepackage{amssymb}
\usepackage{amsthm}
\usepackage{bm}
\usepackage{natbib}  
\usepackage{caption} 
\usepackage{xcolor}
\definecolor{officegreen}{rgb}{0.0, 0.5, 0.0}

\newcommand{\mcL}{\mathcal{L}}

\newcommand{\mbR}{\mathbb{R}}

\newcommand{\mbRd}{{\mathbb{R}^d}}

\def\omg{{\Omega}}

\title{Data-driven learning of nonlocal models:\\ from high-fidelity simulations to constitutive laws}
\author {
    Huaiqian You,\textsuperscript{\rm 1} 
    Yue Yu,\textsuperscript{\rm 1}
    Stewart Silling,\textsuperscript{\rm 2}
    Marta D'Elia\textsuperscript{\rm 3}
     \\
}
\affiliations {
     \textsuperscript{\rm 1} Mathematics Department, Lehigh University, PA\\
    \textsuperscript{\rm 2} Center for Computing Research, Sandia National Laboratories, Albuquerque, NM \\
    \textsuperscript{\rm 3} Computational Science and Analysis, Sandia National Laboratories, Livermore, CA \\
    huy316@lehigh.edu, yuy214@lehigh.edu, sasilli@sandia.gov, mdelia@sandia.gov.
}

\begin{document}

\maketitle

\begin{abstract}
We show that machine learning can improve the accuracy of simulations of stress waves in one-dimensional composite materials.
We propose a data-driven technique to learn nonlocal constitutive laws for stress wave propagation models. The method is an optimization-based technique in which the nonlocal kernel function is approximated via Bernstein polynomials. The kernel, including both its functional form and parameters, is derived so that when used in a nonlocal solver, it generates solutions that closely match high-fidelity data. The optimal kernel therefore acts as a homogenized nonlocal continuum model that accurately reproduces wave motion in a smaller-scale, more detailed model that can include multiple materials. We apply this technique to wave propagation within a heterogeneous bar with a periodic microstructure. Several one-dimensional numerical tests illustrate the accuracy of our algorithm. The optimal kernel is demonstrated to reproduce high-fidelity data for a composite material in applications that are substantially different from the problems used as training data. 
\end{abstract}

\section{Introduction}
Nonlocal models use integral operators acting on a lengthscale $\delta$, known as horizon. This feature allows nonlocal models to capture long-range forces at small scales and multiscale behavior, and to reduce regularity requirements on the solutions, which are allowed to be discontinuous or even singular. In recent decades, nonlocal equations have been successfully used to model several engineering and scientific applications, including fracture mechanics \cite{silling2000reformulation,ha2011characteristics,trask2019asymptotically}, subsurface transport \cite{Benson2000,Schumer2003}, image processing \cite{DElia2019imaging,Gilboa2007},
multiscale and multiphysics systems \cite{Alali2012,Askari2008,you2020asymptotically},
finance \cite{Scalas2000}, and stochastic processes \cite{DElia2017,Meerschaert2012}.

However, it is often the case that {\it nonlocal kernels} defining nonlocal operators are justified a posteriori and it is not clear how to define such kernels to faithfully describe a physical system. The problem of learning an appropriate kernel for a specific application is one of the most challenging open problems in nonlocal modeling. 
The literature on techniques for learning kernel {\it parameters} for a given functional form is vast, see, e.g., \cite{burkovska2020,d2016identification} for control-based approaches, and \cite{pang2020npinns,Pang2019fPINNs} for machine-learning approaches. However, the use of machine learning to learn the {\it functional form} of the kernel is still in its infancy, \cite{Xu2020deriving,Xu2020learning,You2020Regression} being the only relevant works that we are aware of.

In this work we use an approach similar to the one developed in \cite{You2020Regression} to learn nonlocal kernels whose associated nonlocal wave equation is well posed by construction and can be used as an accurate surrogate for more detailed, high-fidelity wave propagation models. In particular, we present an application to wave propagation at the microscale in a heterogeneous solid. In this context, the machine-learned nonlocal kernel embeds the material constitutive behavior so that the material interfaces do not have to be treated explicitly and, more importantly, the material microstructure can be unknown. Furthermore, the corresponding nonlocal models allow for accurate simulations at scales that are much larger than the microstructure.

\noindent Our main contributions are:\\
$\bullet$ The design of an optimization technique that bridges micro and continuum scales by providing accurate and stable model surrogates for the simulation of wave propagation in heterogeneous materials.\\
$\bullet$ The illustration of this method via one-dimensional experiments that confirm the applicability of our technique and the improved accuracy compared with state-of-the-art results.\\
$\bullet$ The demonstration of generalization properties of our algorithm whose associated model surrogates are effective even on problem settings that are substantially different from the ones used for training in terms of loading and time scales.

\section{Nonlocal kernel learning}
We introduce the high-fidelity (HF) model that faithfully represents the system: for $\omg\in\mbRd$, the scalar function $u(x,t)$ solves, for $(x,t) \in \Omega\times[0,T]$
\begin{equation}\label{high-fidelity}
\dfrac{\partial^2 u}{\partial t^2}(x,t)-\mathcal{L}_{\text{HF}} [u](x,t) = f(x,t),
\end{equation}
provided some boundary conditions on $\partial\Omega$ for $u(x,t)$ and initial conditions at $t=0$ for $u$ and $\partial u/\partial t$ are satisfied. Here, $\mcL_{\text{HF}}$ is the HF operator, which can either be a differential or integral operator, and $f$ represents a forcing term.

We assume that solutions to this HF problem may be approximated by solutions to a nonlocal problem of the form
\begin{equation}\label{coarsegrained}
\dfrac{\partial^2 u}{\partial t^2}(x,t)-\mathcal{L}_K[u](x,t) = f(x,t),
\end{equation}
for $(x,t) \in \Omega\times[0,T]$, augmented with nonlocal boundary conditions on $\Omega_\delta$ (a layer of thickness $\delta$ that surrounds the domain) and the same initial conditions on the variable $u$ and its derivative as in \eqref{high-fidelity}. The forcing $f$ may coincide with the forcing term in \eqref{high-fidelity} or it could be an appropriate representation of the same. 

We seek $\mcL_K$ as a nonlocal operator of the form
\begin{equation}\label{LK}
\mcL_K[u](x,t) = \int_{\overline\Omega} K(|x-y|) \left(u(y,t)-u(x,t)\right)\, dy,
\end{equation}
where $K$ is a radial, sign-changing, {\it kernel function}, compactly supported on the ball of radius $\delta$ centered at $x$, i.e., $B_\delta(x)$ and $\overline\Omega=\Omega\cup\Omega_\delta$.

\subsection{The algorithm}
To learn the kernel $K$, we assume that we are given $N$ pairs of forcing terms and corresponding solutions to \eqref{high-fidelity}, normalized with respect to the $L^2$ norm of each solution over $\Omega\times[0,T_{\rm tr}]$. These are denoted by
\begin{equation}
\mathcal{D}_{\rm tr} = \left\{(u_k(x,t),f_k(x,t))\right\}_{k=1}^N,
\end{equation}
for $x\in\Omega$ and $t\in(0,T_{\rm tr}]$.
Similarly to \cite{You2020Regression}, we represent $K$ as a linear combination of Bernstein basis polynomials:
\begin{equation}\label{model-kernel}
\begin{aligned}
K\left(\frac{|y|}{\delta}\right) 
& = \sum_{m=0}^{M}\frac{C_m}{\delta^{d+2}} B_{m,M}
    \bigg (\bigg|\frac{y}{\delta}\bigg|\bigg),
\end{aligned}
\end{equation}
where the Bernstein basis functions are defined as

$B_{m,M}(x) = \begin{pmatrix}
M\\
m\\
\end{pmatrix}
x^m(1-x)^{M-m}\;\;$ for $0\leq x\leq 1$\\ and where $C_m\in\mbR$. Note that, by construction, this kernel guarantees that \eqref{coarsegrained} is well-posed \cite{Du2018peridynamic}.

We machine-learn the nonlocal model by finding optimal parameters $\{C_m\}$ such that solutions $\hat u_k$ to \eqref{coarsegrained}, for $f=f_k$ and the kernel function $K$ associated to $\{C_m\}$, are as close as possible to the training variable $u_k$. 

In this work we numerically approximate $\hat u_k$ by $\bar u_k$ using a central-differencing scheme in time with time step d$t$, i.e.
\begin{equation}\label{ubar}
\begin{aligned}
\bar u_k^{n+1}(x_i)
& = 2\bar u^{n}_k(x_i)- \bar u^{n-1}_k(x_i)\\
&+{\rm d}t^2\left(\mathcal{L}_{K,h }[\bar u^n_k](x_i) + f_k(x_i,t^n)\right),
\end{aligned}
\end{equation}
where $\bar u_k^{n+1}(x_i)$ represents the $k$-th approximate solution at time step $t^{n+1}$ and at discretization point $x_i$, and $\mcL_{K,h}$ is an approximation of $\mcL_K$ by Riemann sum with uniform grid spacing $h$.
The optimal parameters are  obtained by solving the following optimization problem.
\begin{align}
\min_{C_m}& \frac{T_{\rm tr}}{{\rm d}t^3 \,N} \sum_{k=1}^N \sum_{n=1}^{T_{\rm tr}/{\rm d} t} \big\| \bar u_k^{n+1}-u_k(t^{n+1}) \big\|_{\ell^2}^2
+\mathcal R(\{C_m\}),\label{abstractOptProblem}\\
\text{s.t.} & \text{$\;\;\bar u_k$ satisfies  \eqref{ubar} and}
\label{abstract-constraint-state}\\
& \;\;\text{$K$ satisfies physics-based constraints.}\label{abstract-physics}
\end{align}
Here, the $\ell^2$ norm is taken over the space-discretization points $x_i$, $\mathcal R(\cdot)$ is a regularization term on the coefficients that improves the conditioning of the optimization problem, and \eqref{abstract-physics} depends on the physics of the problem (as an example, it may correspond to enforcing that the surrogate model reproduces exactly a certain class of solutions).

\section{Dispersion in heterogeneous materials}
\begin{figure}[t]
\centering
\includegraphics[width=0.99\columnwidth]{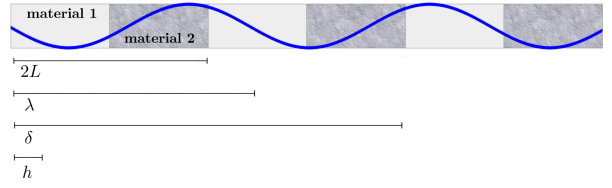}
\caption{One-dimensional bar with ordered microstructure of period $2L$. Material 1 and 2 have the same density and Young modulus $E_1$ and $E_2$. The horizon $\delta$, the wave length $\lambda$, and the discretization size, $h$, are reported for comparison.}
\label{fig:bar}
\end{figure}

We apply the learning algorithm described above to the propagation of waves in a one-dimensional heterogeneous bar, like the one reported in Figure \ref{fig:bar}, with an {\it ordered} microstructure, i.e. two materials with the same length alternate periodically. Our goal is to learn a nonlocal model able to reproduce wave propagation on distances that are much larger than the size of the microstructure without resolving the microscales.  
The high-fidelity model we rely on is the classical wave equation; the corresponding high-fidelity data used for training and validation are obtained with the solver described below. 

\subsection{High-fidelity data}
\begin{figure}  [t!]
\centering
\includegraphics[width=0.5\textwidth]{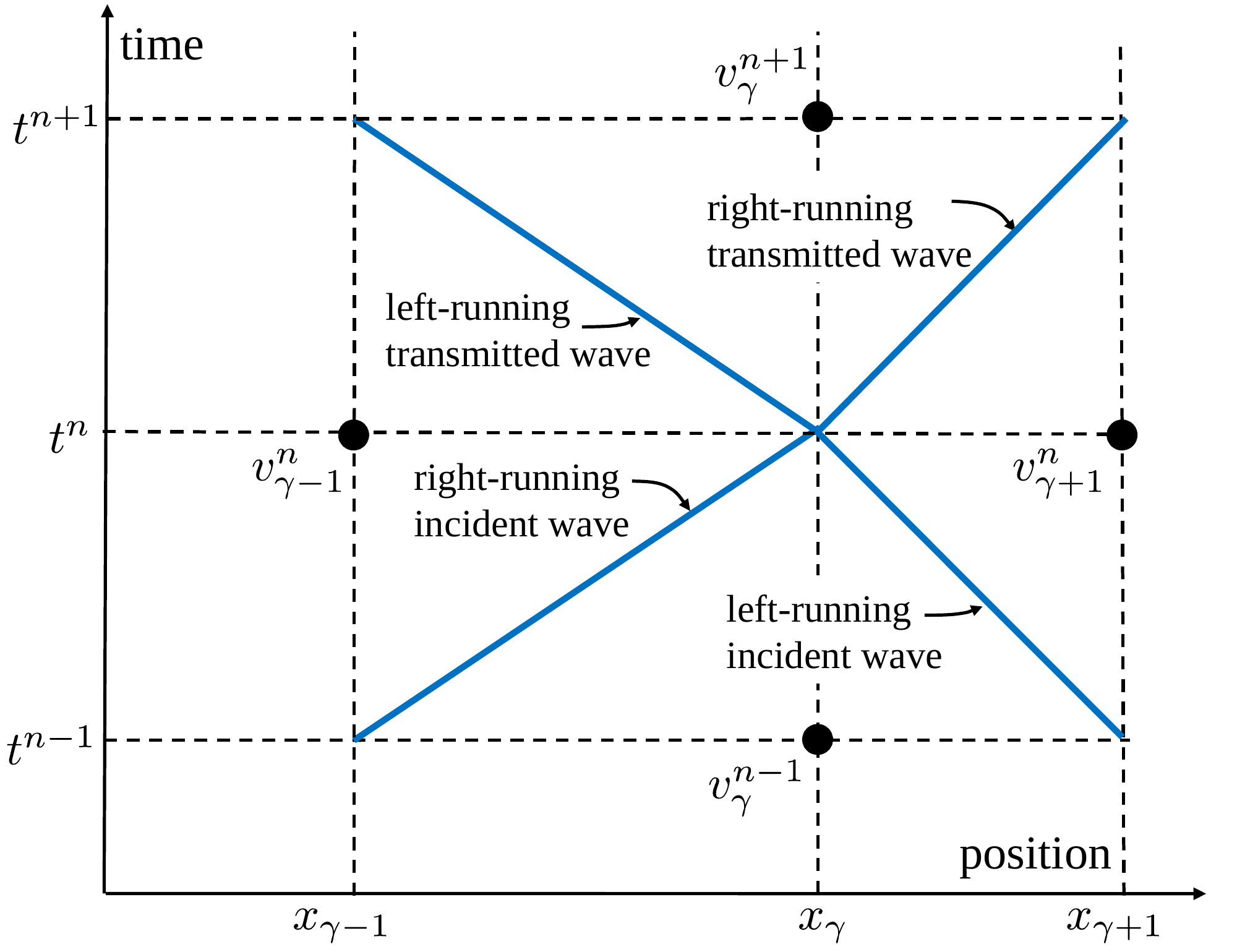}
\caption[Interaction of two elastic waves in the DNS method.]
{Interaction of two waves in the DNS method. 
Each node may or may not be located at a material interface.}
\label{fig-dns}
\end{figure}

For both training and validation purposes we generate data using high-fidelity simulations for the propagation of stress waves within the microstructure of the heterogeneous, linear elastic bar.
This method, which will be referred to as Direct Numerical Solution (DNS),
constructs an arbitrarily complex wave diagram (also called an $x$-$t$ diagram), that
treats the mutual interaction and superposition of many wavefronts moving in either direction.
The bar is discretized into nodes such that it takes a constant amount of time $\Delta t$
for a wave to travel between nodes $\gamma$ and $\gamma+1$, regardless of the elastic wave speed
in the material between these two nodes.
Therefore, in a heterogeneous medium, the spacing between nodes is not constant. Each node $\gamma$, at each time step $n$, has velocity $v_\gamma^n$ (note that, in this case, the subscript refers to position, as opposed to the previous section where it corresponds to a specific sample $k$).
To compute the velocities in the next time step, it is assumed that two waves moving in
opposite directions converge on the node $\gamma$ at time step $n$ (see Figure~\ref{fig-dns}).
The waves shown in the figure can have unequal slopes on the $x$-$t$ diagram because the
materials on either side of node $\gamma$ can have different waves speeds $c$.
The jump conditions for the waves are applied that relate the stress change $[\sigma]$ across a
wave to the velocity change $[v]$.
These jump conditions have the following form:
\begin{equation*}
   [\sigma]=\pm\rho c [v],
\label{eqn-dns-jump}
\end{equation*}
where $\rho$ is the mass density, and
where the $+$ and $-$ signs apply to right-running waves and left-running waves respectively.
From these conditions, the velocity $v_\gamma^{n+1}$ can be computed explicitly from the values
at the adjacent nodes in time step $n-1$.
Externally applied forces can also be included in a straightforward way. After $v_\gamma^{n+1}$ is computed, the updated displacement is approximated by
\[    u_\gamma^{n+1} = u_\gamma^n + \Delta t v_\gamma^{n+1}.   \]
Details of the method can be found in \cite{Silling2020Pulse}.

This DNS solver has the important advantage of not using an approximate representation of derivatives in space or time for the computation of the velocity, which is, therefore, free from truncation error and other sources of discretization error that
are usually encountered with PDE solvers.
This allows us to model the propagation of waves through many thousands of microstructural interfaces without the need to worry about what features of the velocity are real and what
are numerical artifacts.

\smallskip
We consider four types of data and use the first two for training and the last two for validation of our algorithm. In all our experiments we set $L\!=\!0.2$, $E_1\!=\!1$, $E_2\!=\!0.25$, $\rho\!=\!1$, and the symmetric domain $\Omega\!=\!(-b,b)$. Discretization parameters for the DNS solver are set to $\Delta t\!=\!0.01$ and $\max\{\Delta x\}\!=\!0.01$.

\medskip\noindent
{\it 1) Oscillating source.} We set $b=50$, $v(x,0)=u(x,0)=0$,

\noindent $f(x,t) \!= e^{-\left(\frac{2x}{5kL}\right)^2} \!e^{-\left(\frac{t-t_0}{t_p}\right)^2}\!\cos^2\left(\frac{2\pi x}{kL}\right), \, k=1,2,\ldots,20,$ $t_0=t_p=0.8$.

\medskip\noindent
{\it 2) Plane wave.} For $b=50$, $f(x,t)=0$ and $u(x,0)=0$, we prescribe  
$v(-b,t)=\sin(\omega t)$ for $\omega=0.35,0.7,\cdots,3.85$.

\medskip\noindent
{\it 3) Wave packet.} For $b=133.3$, $f(x,t)=0$ and $u(x,0)=0$, we prescribe 
$v(-b,t)=\sin(\omega t)\exp(-(t/5-3)^2)$ for $\omega=2,3.9,5$.

\medskip\noindent
{\it 4) Impact.} For $b=$266.6, $f(x,t)=0$ and $u(x,0)=0$, we prescribe $v(x,0)=1$ for all $x\in [-b,-b+1.6]$ and $v=0$ outside of this interval. This initial condition represents an impactor hitting the bar at time zero, generating a velocity pulse of width roughly 3.2 that propagates into the interior  of the bar. The pulse attenuates and changes shape as it encounters the many microstructural interfaces.

\subsection{Training procedure}
For the optimization problem \eqref{abstractOptProblem} we choose a Tikhonov regularization of the form
$\mathcal R(\{C_m\})=\frac{\epsilon}{M+1}\sum_{m=0}^M C_m^2$, where the regularization weight $\epsilon$ is chosen empirically to guarantee accurate predictions, as we explain later on. The physics-based constraints in \eqref{abstract-physics} are defined as follows and also discretized by Riemann sum; they are used to explicitly prescribe values of $C_{M-1}$ and $C_M$:
\begin{equation}\label{physics-based-c}
\begin{aligned}
\sum_{m=0}^{M}C_m & \int_0^{\delta}\frac{ y^2}{\delta^{3}} B_{m,M} \bigg (\frac{|y|}{\delta}\bigg)dy=\rho c^2_0,\\
\sum_{m=0}^{M}C_m & \int_0^{\delta}\frac{y^4}{\delta^{3}} B_{m,M} \bigg (\frac{|y|}{\delta}\bigg)dy=-4\rho c_0^3R,
\end{aligned}
\end{equation}
where $\rho$ is the density and $c_0$ is the effective wave speed for infinitely long wavelengths. 
For $\rho=1$, it is given by 
$c_0=(2/(1/E_1+1/E_2)^\frac12$.
$R$ is the second derivative of the wave group velocity with respect to the frequency $\omega$ evaluated at $\omega=0$. 
Both parameters are obtained by simulating a very low frequency plane wave propagating through the microstructure over a long distance using DNS \cite{Silling2020Pulse}.
These parameters primarily affect simulations at large times, $t>10$.
However, due to practical limitations on computer resources, our training simulations are restricted to $t\le2$. Therefore, we incorporate these parameters as constraints obtained from DNS as indicated in \eqref{physics-based-c}, rather than attempting to learn these through our algorithm.
The first constraint in \eqref{physics-based-c} is also used for similar purposes in \cite{Xu2020learning} and prescribed in a weak sense by penalization. 

Training is performed with DNS data of type 1) and 2). Parameters for the nonlocal solver and the optimization algorithm are set to $h=0.05$, ${\rm d} t=0.02$, $T_{\rm tr}=2$, $\delta=$1.2, $M=24$ and $\epsilon=0.01$. The optimization problem \eqref{abstractOptProblem} is solved with L-BFGS. Note that we empirically choose $\delta$ and $\epsilon$ in such a way that the group velocity, defined below, corresponding to the optimal kernel is as close as possible to the one computed with DNS.

\begin{figure}[t]
\centering
\includegraphics[width=\columnwidth]{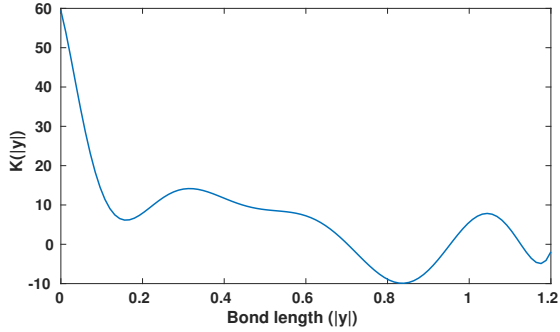}
\caption{Optimal kernel $K_{\rm opt}$ as a function of distance.}
\label{fig:opt-kernel}
\end{figure}

The optimal kernel, $K_{\rm opt}$, is reported in Figure \ref{fig:opt-kernel}; as expected from the literature \cite{Xu2020deriving,Xu2020learning,You2020Regression,weckner2011determination}, we observe a sign-changing behavior. 
We also compute the corresponding dispersion $\omega(k)$ and group velocity $v_g(\omega)=d\omega/dk$. For a given kernel $K$ and different frequencies $k_i=0,\frac{2\pi}{200h},\cdots,\frac{2\pi}{h}$, the corresponding angular frequency $\omega(k_i)$ and group velocity $v_g(\omega(k_i))$ are approximated by
\begin{displaymath}
\begin{aligned}
\omega(k_i)^2&\approx\frac{1}{\rho}\sum_{q}K(|y_q|)(1-\cos(k_i y_q))h,\\
v_g(\omega(k_i))&\approx\dfrac{\omega(k_{i+1})-\omega(k_{i-1})}{k_{i+1}-k_{i-1}},
\end{aligned}
\end{displaymath}
where $y_q$ belong to a uniform grid of size $h$ in $(-\delta,\delta)$. 

\begin{figure}[t]
\centering
\includegraphics[width=\columnwidth]{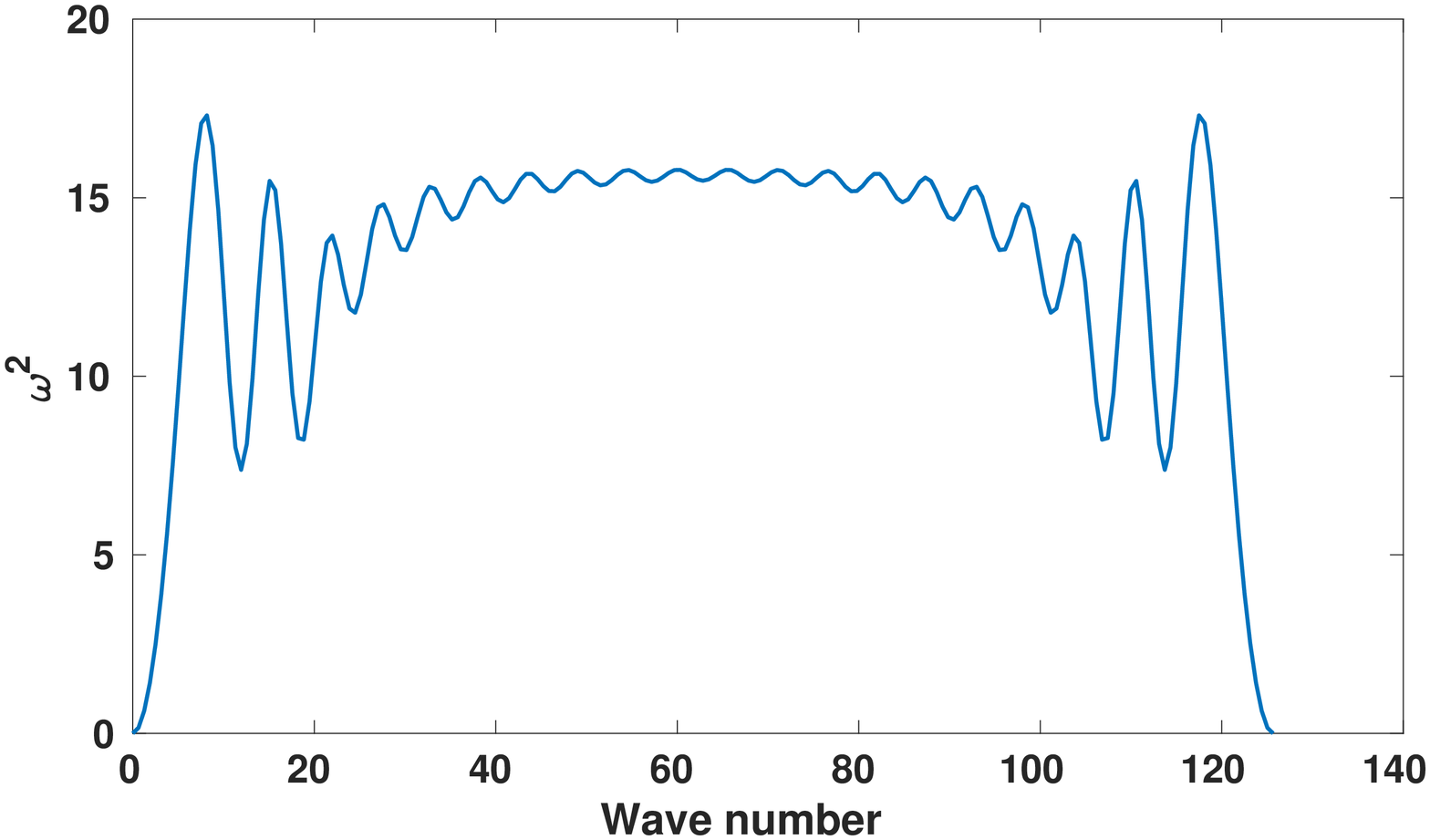}
\caption{Dispersion curve associated with $K_{\rm opt}$.}
\label{fig:dispersion}
\end{figure}

\begin{figure}[t]
\centering
\includegraphics[width=\columnwidth]{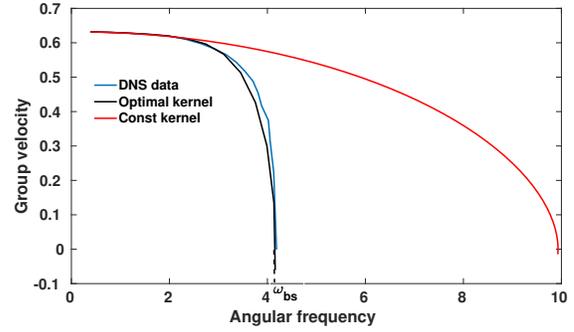}
\includegraphics[width=\columnwidth]{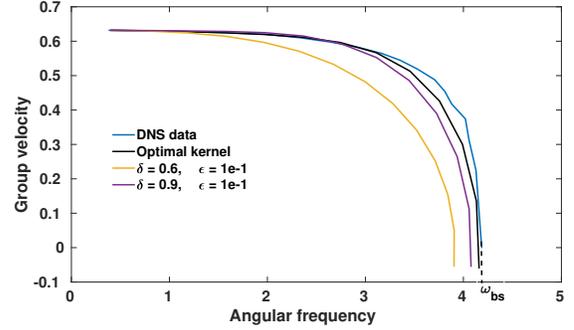}
\caption{Comparison of the group velocity. Upper: group velocity corresponding to $K_{\rm opt}$, $K_{\rm const}$, and DNS data. Bottom: group velocity corresponding to $K_{\rm opt}$ for different pairs $(\delta,\epsilon)$.}
\label{fig:group-vel}
\end{figure}

The dispersion curve is reported in Figure \ref{fig:dispersion}, its positivity indicates that $K_{\rm opt}$ corresponds to a physically stable material model. 
The group velocity is reported in the upper plot of Figure \ref{fig:group-vel} in comparison with the curve computed with DNS by observing the speed of a wave packet of a given frequency as it moves through the microstructure. We also display the group velocity associated with an alternative kernel obtained for the same material by a completely different method \cite{Silling2020Pulse}. This alternative kernel is a constant, specifically, we have that $C_m\!=\!K_{\rm const}\!=\!0.7714$, for $M\!=\!3$ and $\delta\!=\!0.15$.

It is well known that layered, periodic elastic media have a band structure for wave propagation, see \cite[pages 121-122]{Bedford1994}. In the present study, because it is not possible to reproduce the higher-frequency pass bands with the coarse discretization that is used in fitting our nonlocal kernel, we address only the first, low-frequency pass band, i.e. $\omega\in(0,\omega_{\rm bs})$, where ``bs'' stands for band stop. Hence, the optimal kernel is suitable only for wavelengths that are bigger than the microstructure; this is enough to reproduce the physically most important features of wave propagation in layered media for typical applications. 

The profile of the group velocity shows the improved accuracy of our optimal kernel that not only matches the behavior for low values of $\omega$, but also catches the behavior at $\omega=\omega_{\rm bs}\approx 4$. This fact has important consequences on the ability to reproduce wave propagation for values of $\omega$ bigger than $\omega_{\rm bs}$. In order to justify the statement above on the optimality of the parameters $\delta$ and $\epsilon$, we report the group velocity profile in correspondence of different pairs; it is clear from the profiles in Figure \ref{fig:group-vel} that $(\delta,\epsilon)=(1.2,0.01)$ provides the best match both in terms of curvature at $\omega=0$ and identification of the band stop.

\begin{figure}[t]
\centering
\includegraphics[width=1\columnwidth]{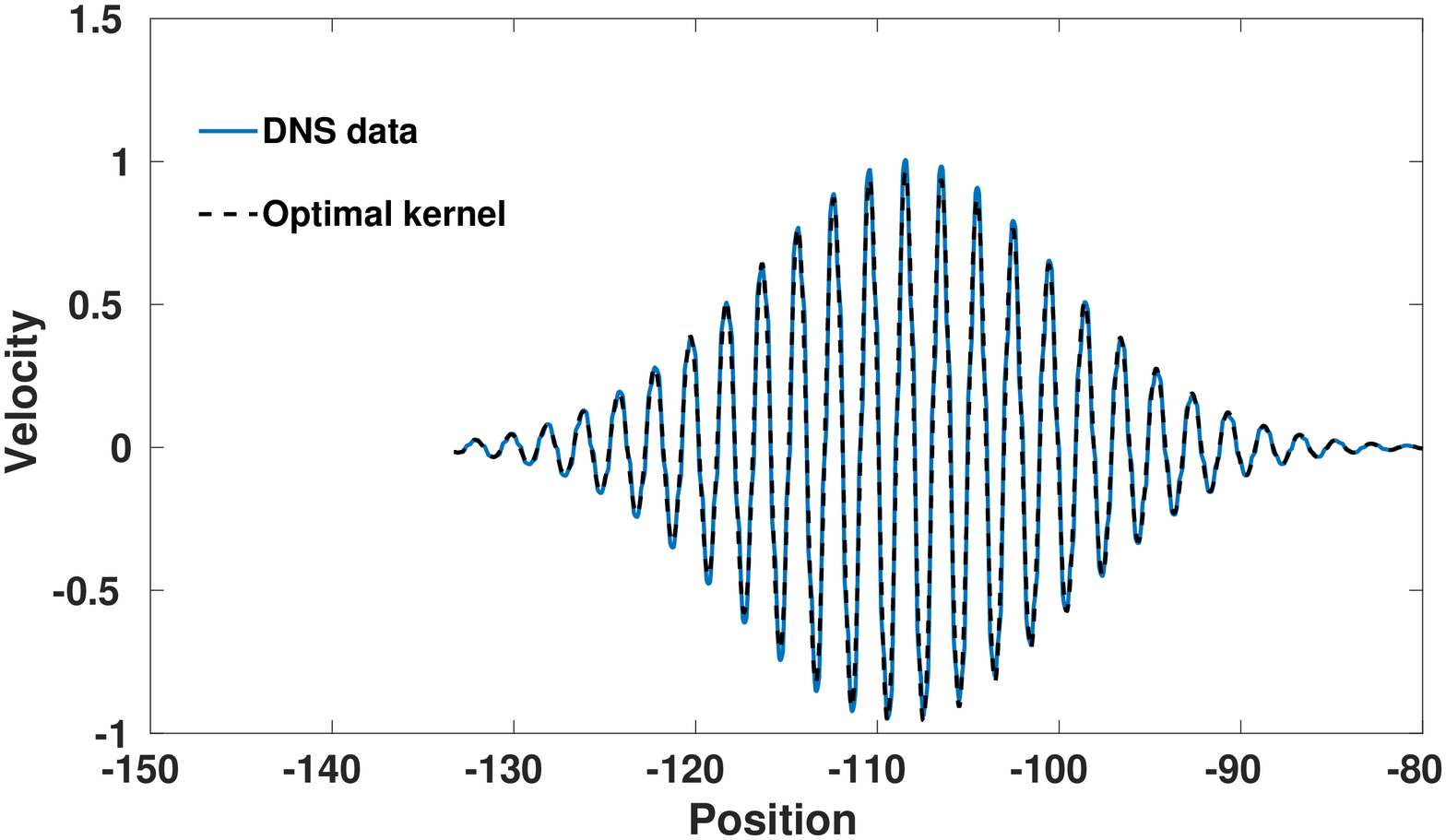}\vspace{-0.12in}
\includegraphics[width=1\columnwidth]{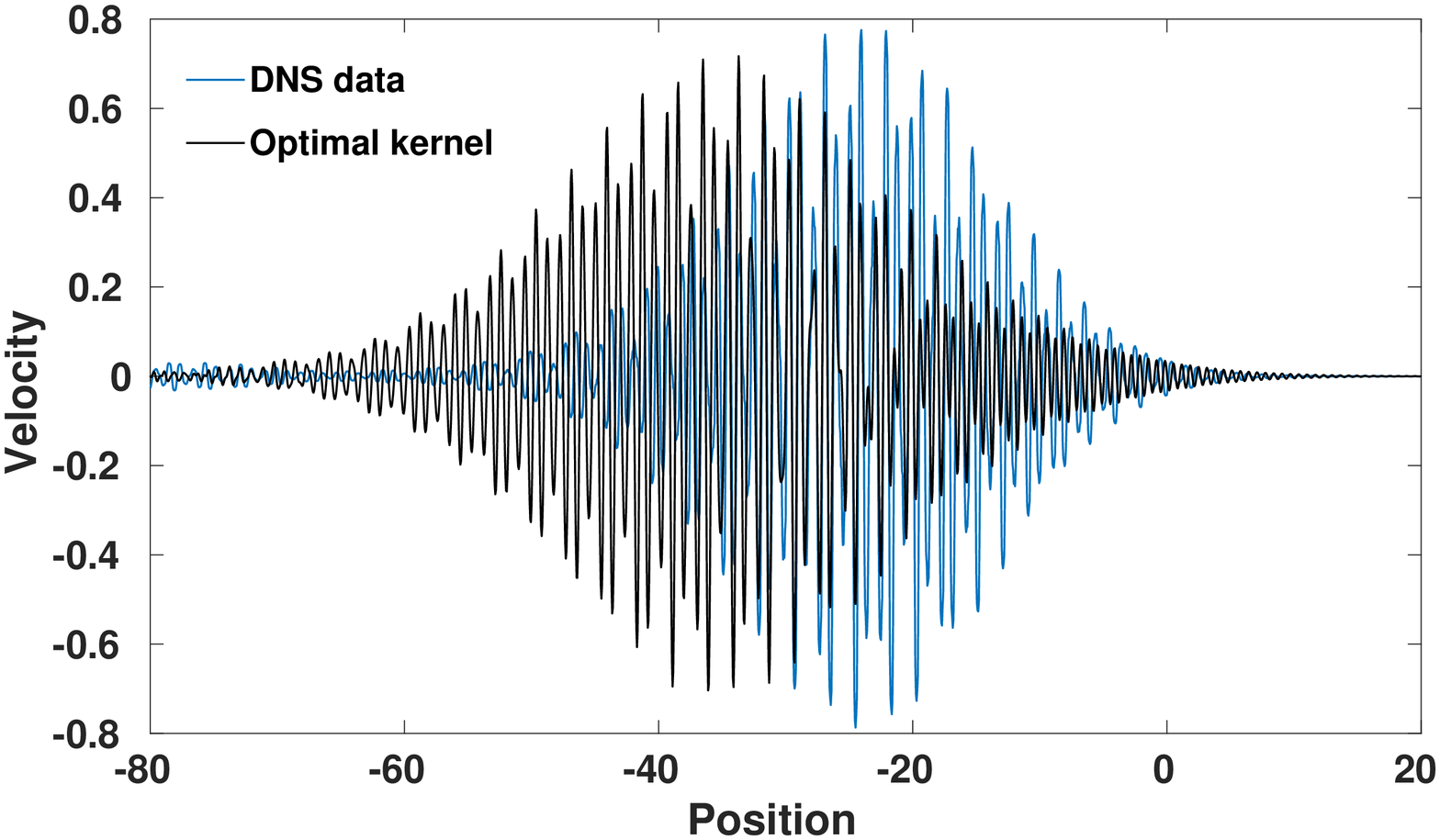}\vspace{-0.12in}
\includegraphics[width=1\columnwidth]{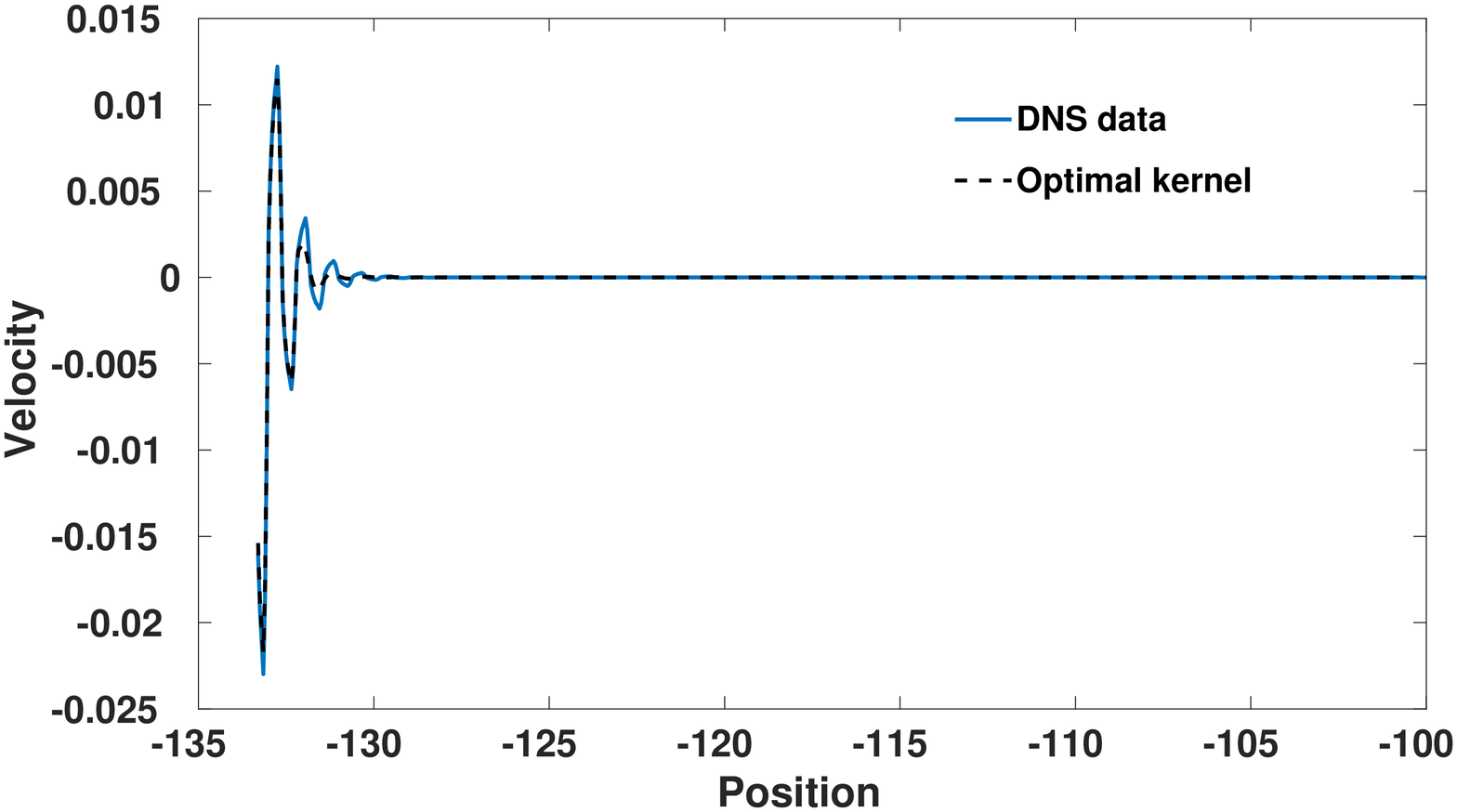}
\caption{Velocity computed with $K_{\rm opt}$. Plots from top to bottom correspond to: 1. $\omega_1=2$ at $t=100$; 2. $\omega_2=3.9$ at $t=320$; 3. $\omega_3=5$ at $t=100$.}
\label{fig:packet-omega1-vel}
\end{figure}



\begin{figure}[t]
\centering
\includegraphics[width=1\columnwidth]{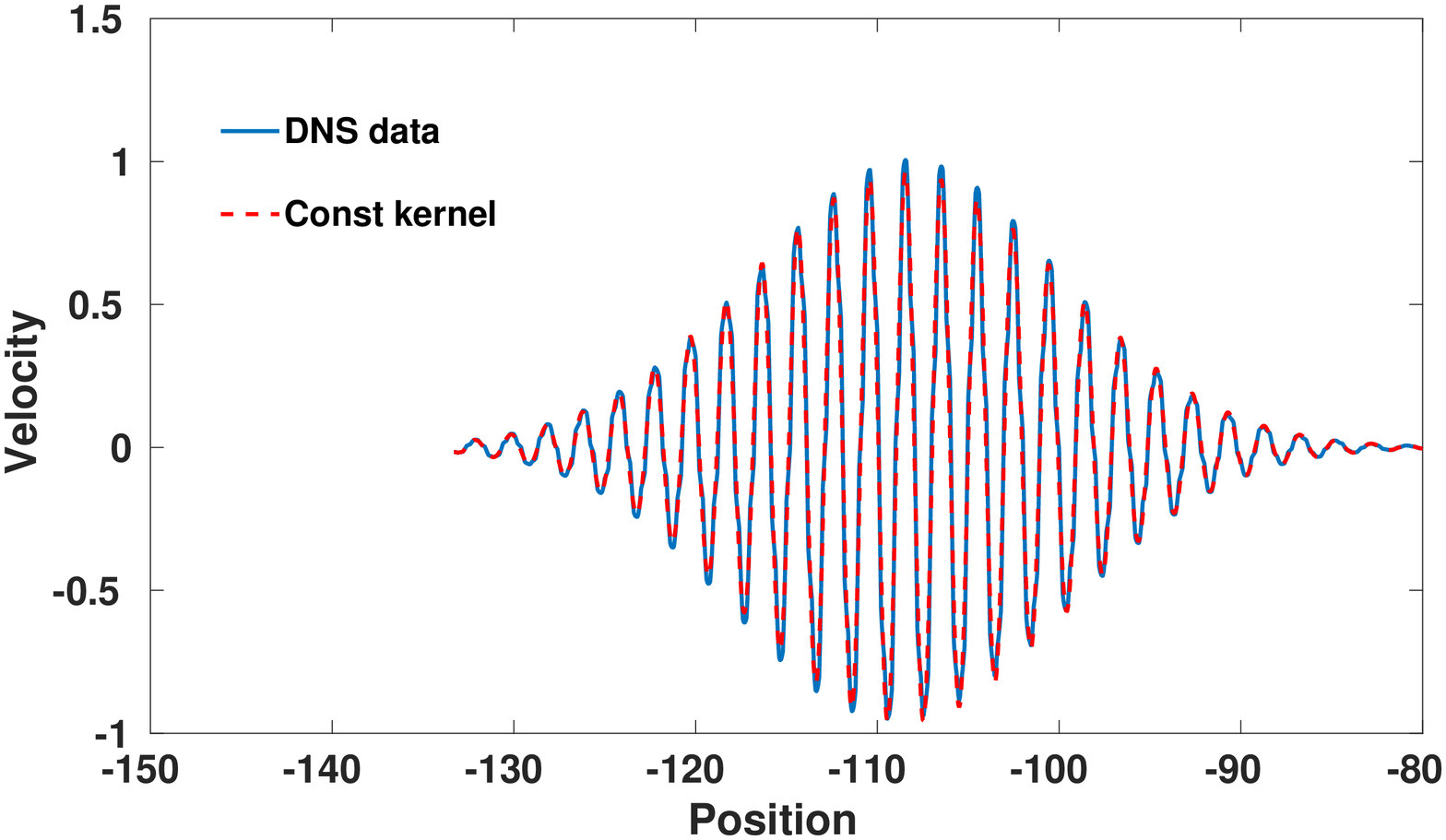}\vspace{-0.12in}
\includegraphics[width=1\columnwidth]{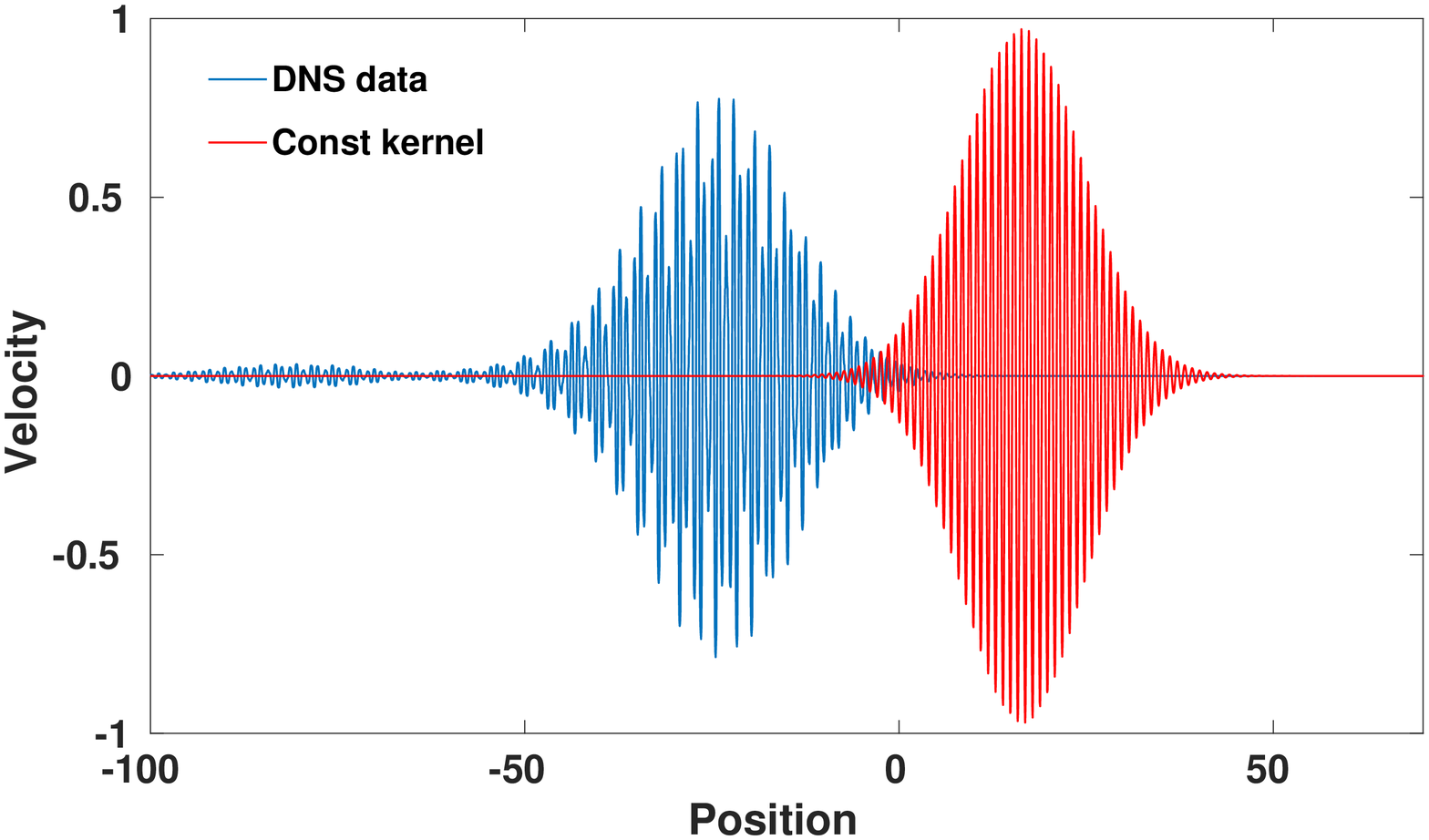}\vspace{-0.12in}
\includegraphics[width=1\columnwidth]{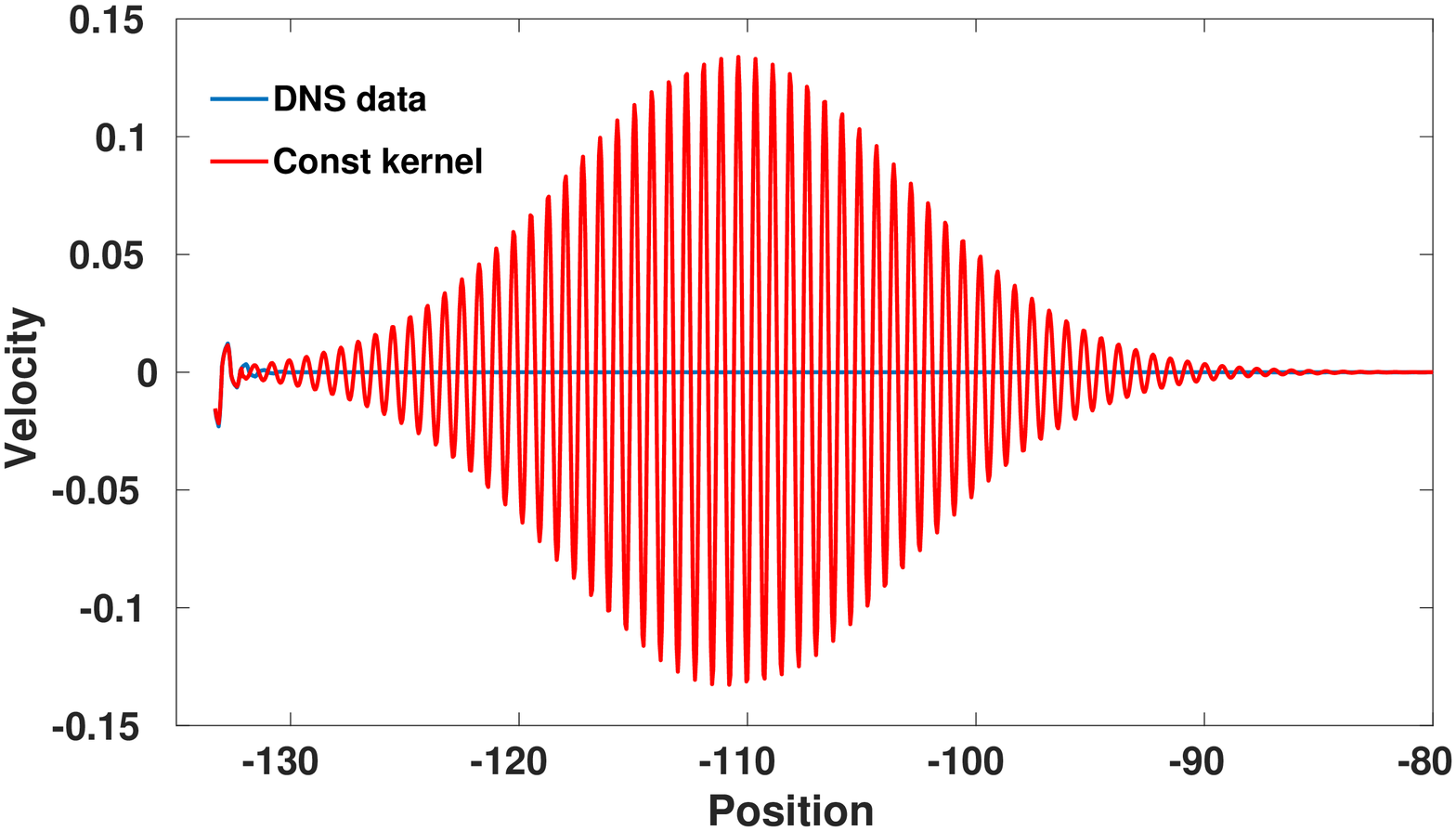}
\caption{Velocity computed with $K_{\rm const}$. Plots from top to bottom correspond to: 1. $\omega_1=2$ at $t=100$; 2. $\omega_2=3.9$ at $t=320$; 3. $\omega_3=5$ at $t=100$.
The optimal kernel $K_{\rm opt}$ obtained by machine learning (Figure~\ref{fig:packet-omega1-vel}) clearly performs better than $K_{\rm const}$ for the second and third cases ($\omega_2$ and $\omega_3$).
}
\label{fig:packet-omega2-const}
\end{figure}





\subsection{Numerical validation}
We test the performance of the optimal kernel $K_{\rm opt}$ on data sets of type 3) and 4), i.e. the problem setting considered for validation has different model parameters, including the domain, than the one used for training and, hence, these tests serve as an indicator of the generalization properties of our algorithm. 

\medskip\noindent{\it Wave packet.} For data type 3) we numerically compute solutions to \eqref{coarsegrained} using $K_{\rm opt}$ and DNS data as nonlocal boundary conditions. We consider solutions corresponding to three values of $\omega$: $\omega_1\!=\!2<\!\omega_{\rm bs}$, $\omega_2\!=\!3.9\approx\omega_{\rm bs}$ and $\omega_3\!=\!5>\!\omega_{\rm bs}$. Note that the latter value is beyond the band stop and, as such, corresponds to a zero group velocity, i.e. the wave does not travel in time. In Figure \ref{fig:packet-omega1-vel} we report the velocity corresponding to the computed displacement $\bar u$ at time $t=100$, $t=320$, and $t=100$ for $\omega_1$, $\omega_2$, and $\omega_3$ respectively; as a reference, we also report the exact DNS velocity.
Our results indicate that our kernel can accurately reproduce solutions of type 3) at times larger than $T_{\rm tr}$ and for all values of $\omega$, even larger than $\omega_{\rm bs}$. This is possible because the group velocity corresponding to $K_{\rm opt}$ reproduces the true group velocity very accurately, see Figure \ref{fig:group-vel}. In particular, detecting the presence of a band stop allows us to accurately predict the wave propagation for values of $\omega\!>\!\omega_{\rm bs}$. Due to the poor accuracy of the group velocity associated to $K_{\rm const}$, corresponding solutions are not as accurate for $\omega$ in the proximity of $\omega_{\rm bs}$ and beyond. To illustrate this phenomenon, we report in Figure \ref{fig:packet-omega2-const} the behavior of the velocity corresponding to $K_{\rm const}$ at time $t=100$, $t=320$ and $100$, respectively for $\omega_1$, $\omega_2$ and $\omega_3$. Comparison with DNS data shows that, for $\omega_2$ the wave associated with $K_{\rm const}$ is traveling faster than the exact one and, for $\omega_3$, it keeps traveling while the exact wave is not propagating.

\begin{figure}[t]
\centering
\includegraphics[width=1\columnwidth]{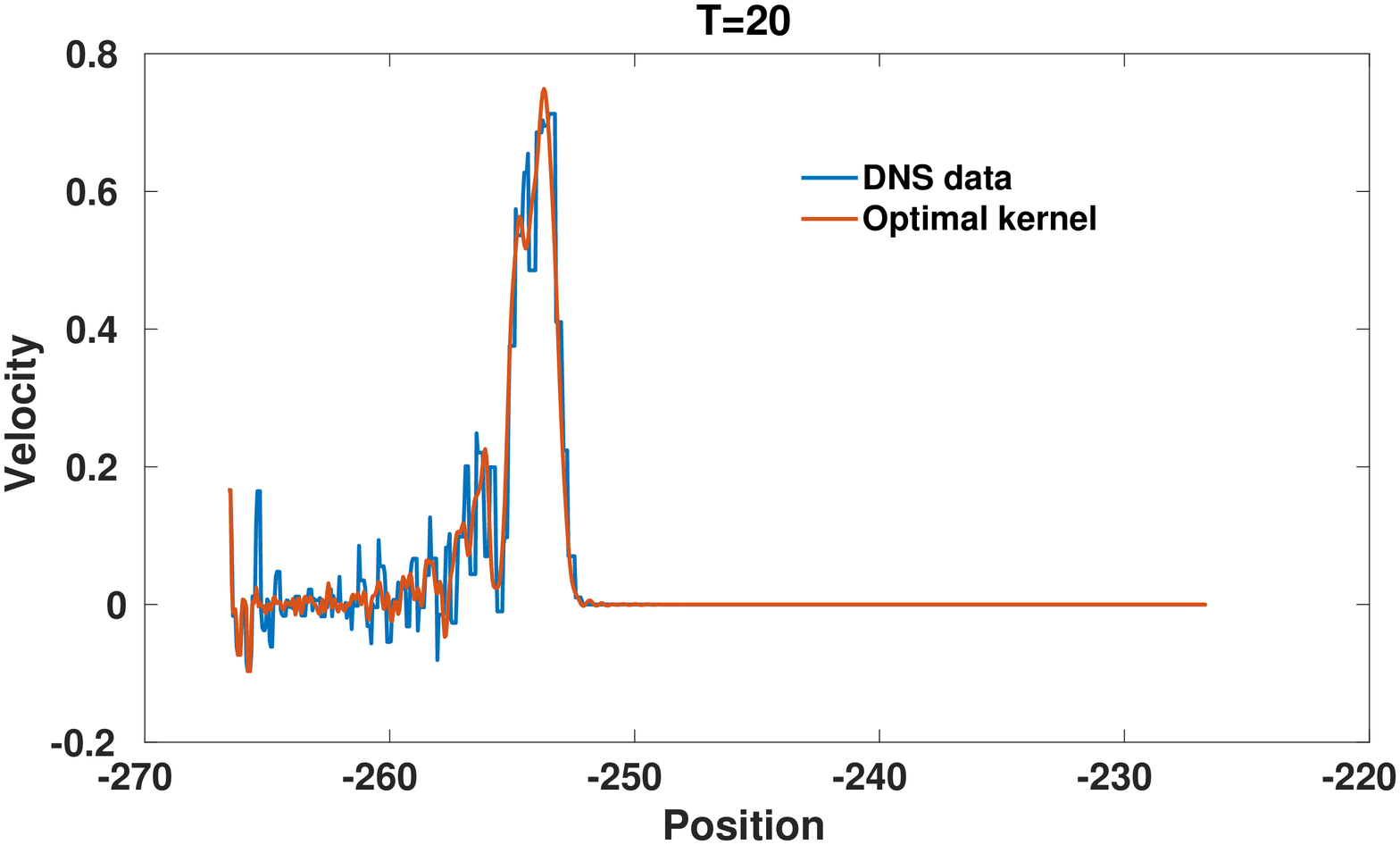}\vspace{-0.11in}
\includegraphics[width=1\columnwidth]{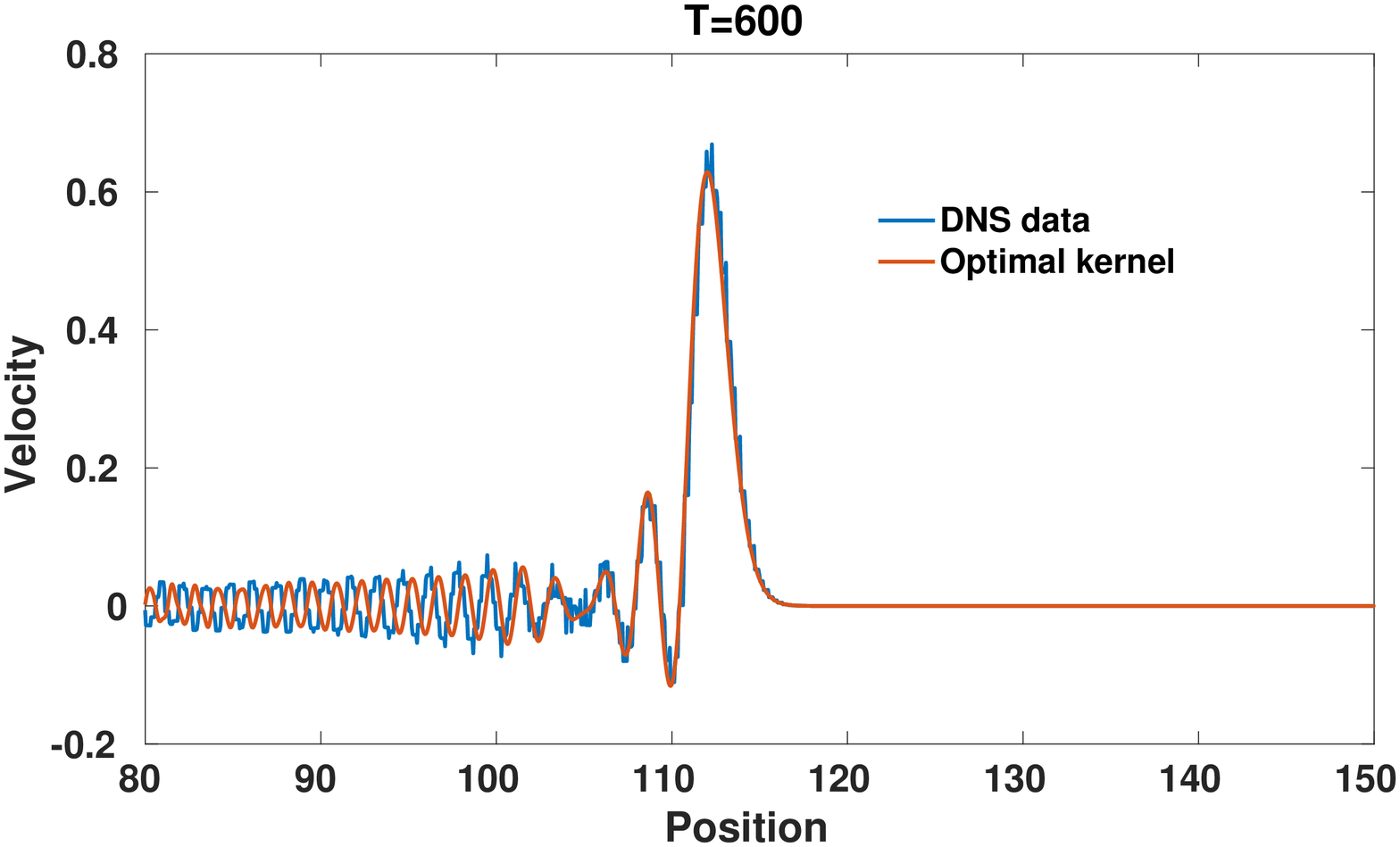}
\caption{Velocity profile for the {\it Impact} problem at $T=20$ and $T=600$ with $K_{\rm opt}$.}
\label{fig:impact-t1-opt}
\end{figure}

\begin{figure}[t]
\centering
\includegraphics[width=1\columnwidth]{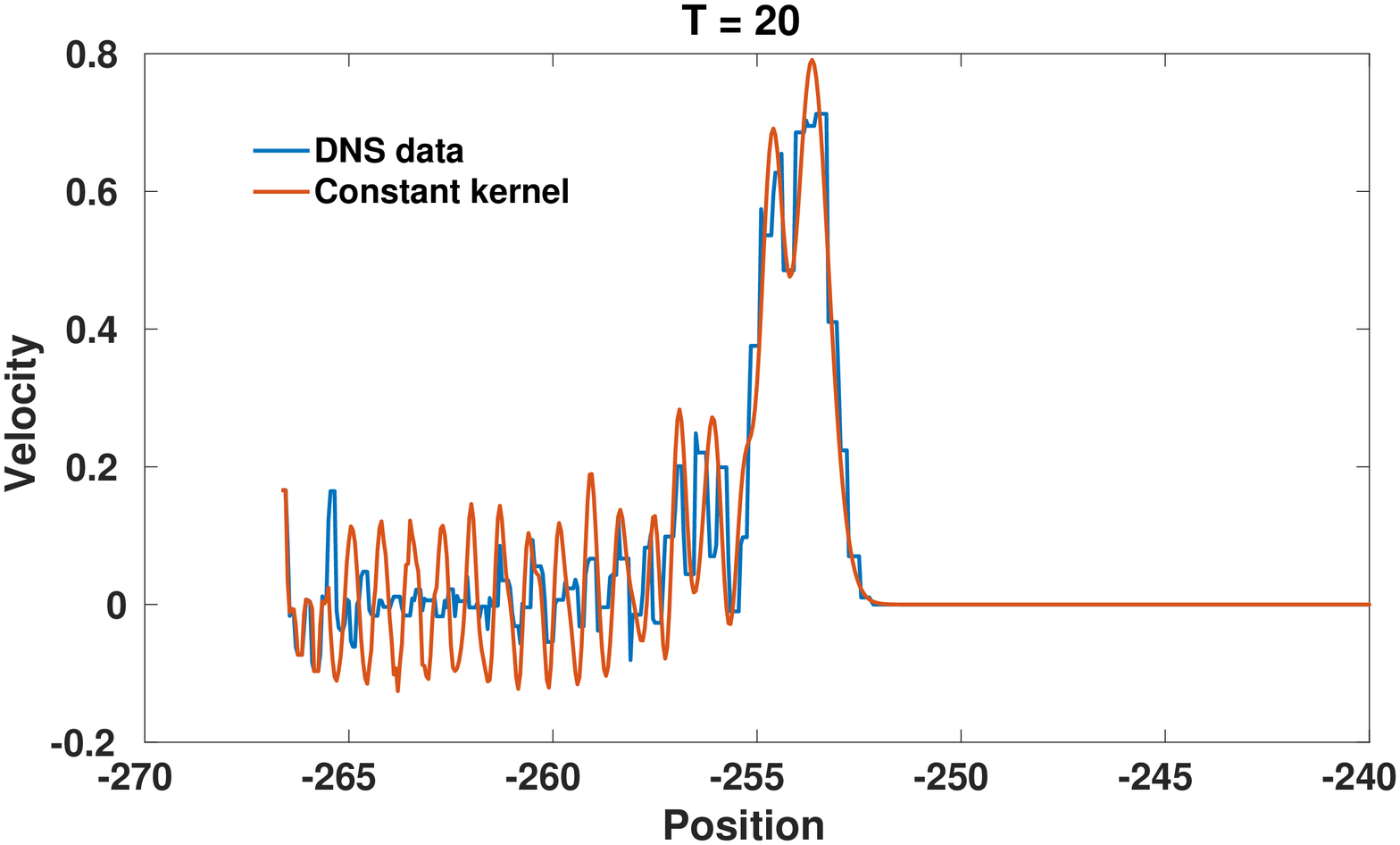}\vspace{-0.11in}
\includegraphics[width=1\columnwidth]{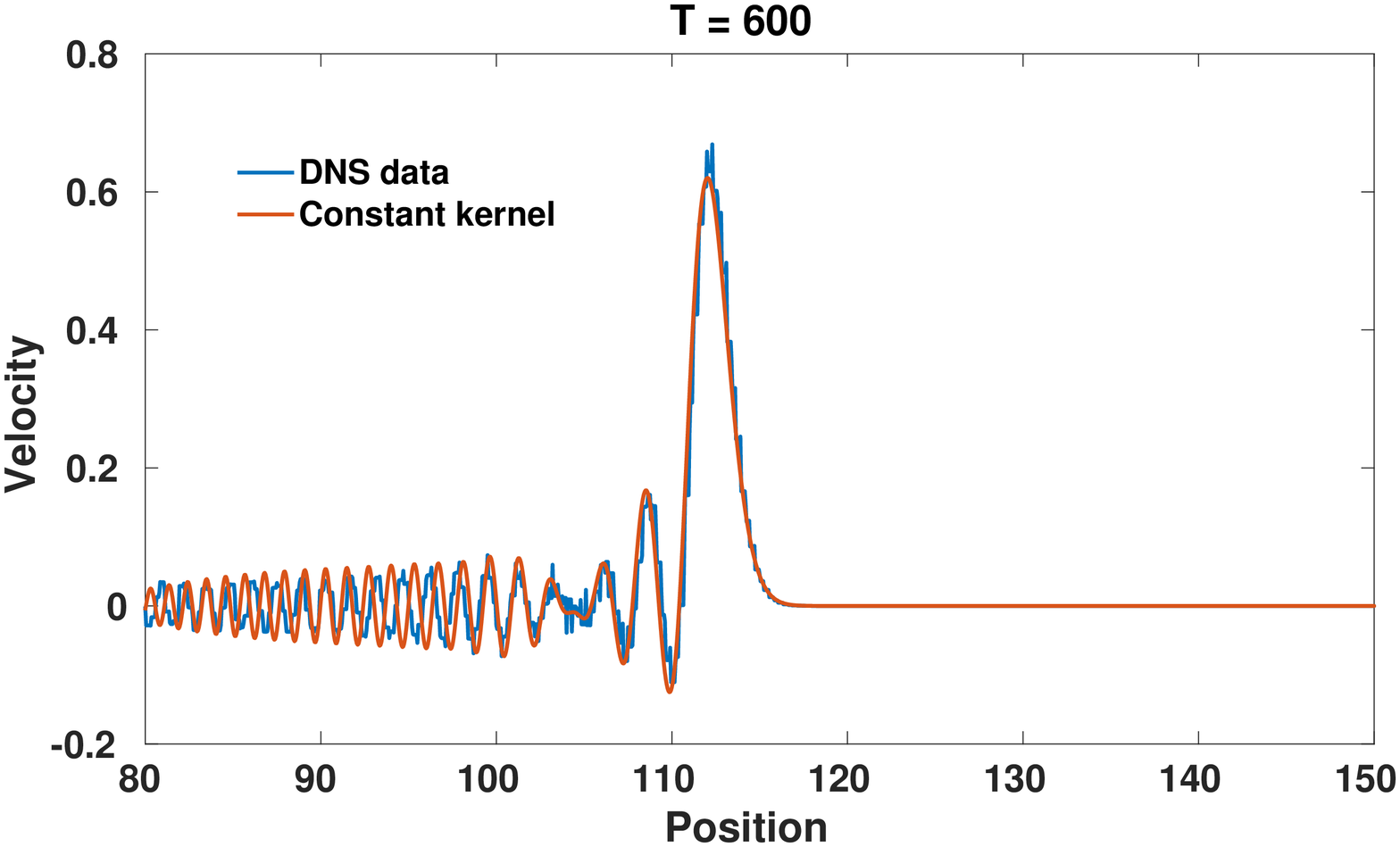}
\caption{Velocity profile for the {\it Impact} problem at $T=20$ and $T=600$ with $K_{\rm const}$.
The optimal kernel $K_{\rm opt}$ obtained by machine learning (Figure~\ref{fig:impact-t1-opt}) provides better agreement with the DNS data than 
$K_{\rm const}$ at the earlier time ($T=20$) by reducing the size of the oscillations that trail the main pulse.
For large $T$, the solution is dominated by the low frequency components of the pulse, for which the two kernels behave similarly.}
\label{fig:impact-t1-const}
\end{figure}

\medskip\noindent{\it Impact.} We use the optimal kernel to compute solutions corresponding to data type 4). In Figure \ref{fig:impact-t1-opt} we report the velocity profile at different time steps in correspondence of $K_{\rm opt}$ and DNS data, displayed for comparison.  Figure \ref{fig:impact-t1-const} displays the same results in correspondence of $K_{\rm const}$.
These results indicate that our optimal kernel can accurately predict the short- and long-time wave propagation, as opposed to the constant kernel that successfully predicts the long-time behavior only. We also point out that for values of $(\delta,\epsilon)$ for which the group velocity is not accurate, the predicted velocity and displacement exhibit non-physical oscillations, which disappear in correspondence of pairs that guarantee an accurate group velocity profile.

\section{Conclusion}
We introduced a new data-driven, optimization-based algorithm for the identification of nonlocal kernels in the context of wave propagation through material featuring heterogeneities at the microscale. The corresponding nonlocal model is well-posed by construction and allows for accurate simulations at a larger scale than the microstructure.
We stress the fact that our algorithm does not require a priori knowledge of the microstructure (often unknown and/or hard to model), but only requires high-fidelity measurements of the displacements or the velocity.
We also point out that our algorithm has excellent generalization properties as the optimal kernel performs well at much larger times than the time instants used for training and on problem settings that are substantially different from the training data set. 

One of the most important findings in this work is the key role of the group velocity in the accuracy of the predictions; in fact, our criterion for the choice of the horizon $\delta$ and the regularization weight $\epsilon$ is the accurate prediction of the group velocity profile. Given the critical role of such quantity, our future work includes the identification of the optimal horizon by, possibly, embedding constraints on the group velocity to the training procedure. Another natural follow-up work is the illustration of the efficiency of our algorithm on two- and three-dimensional test cases.

\section{Acknowledgments}
MD and SS are supported by the Sandia National Laboratories (SNL) Laboratory-directed Research and Development program and by the U.S. Department of Energy, Office of Advanced Scientific Computing Research under the Collaboratory on Mathematics and Physics-Informed Learning Machines for Multiscale and Multiphysics Problems (PhILMs) project. SNL is a multimission laboratory managed and operated by National Technology and Engineering Solutions of Sandia, LLC., a wholly owned subsidiary of Honeywell International, Inc., for the U.S. Department of Energy's National Nuclear Security Administration under contract DE-NA-0003525. This paper, SAND2020-13633, describes objective technical results and analysis. Any subjective views or opinions that might be expressed in this paper do not necessarily represent the views of the U.S. Department of Energy or the United States Government. 
HY and YY are supported by the National Science Foundation under award DMS 1753031.


\bibliography{references}

\begin{thebibliography}{24}
\providecommand{\natexlab}[1]{#1}
\providecommand{\url}[1]{\texttt{#1}}
\providecommand{\urlprefix}{URL }
\expandafter\ifx\csname urlstyle\endcsname\relax
  \providecommand{\doi}[1]{doi:\discretionary{}{}{}#1}\else
  \providecommand{\doi}{doi:\discretionary{}{}{}\begingroup
  \urlstyle{rm}\Url}\fi

\bibitem[{Alali and Lipton(2012)}]{Alali2012}
Alali, B.; and Lipton, R. 2012.
\newblock Multiscale dynamics of heterogeneous media in the peridynamic
  formulation.
\newblock \emph{Journal of Elasticity} 106(1): 71--103.

\bibitem[{Askari et~al.(2008)Askari, Bobaru, Lehoucq, Parks, Silling, and
  Weckner}]{Askari2008}
Askari, E.; Bobaru, F.; Lehoucq, R.; Parks, M.~L.; Silling, S.~A.; and Weckner,
  O. 2008.
\newblock Peridynamics for multiscale materials modeling.
\newblock In \emph{Journal of Physics: Conference Series}, volume 125, 012078.
  IOP Publishing.

\bibitem[{Bedford and Drumheller(1994)}]{Bedford1994}
Bedford, A.; and Drumheller, D.~S. 1994.
\newblock \emph{Introduction to Elastic Wave Propagation}.
\newblock Wiley.

\bibitem[{Benson, Wheatcraft, and Meerschaert(2000)}]{Benson2000}
Benson, D.; Wheatcraft, S.; and Meerschaert, M. 2000.
\newblock Application of a fractional advection-dispersion equation.
\newblock \emph{Water Resources Research} 36(6): 1403--1412.

\bibitem[{Burkovska, Glusa, and D'Elia(2020)}]{burkovska2020}
Burkovska, O.; Glusa, C.; and D'Elia, M. 2020.
\newblock An optimization-based approach to parameter learning for fractional
  type nonlocal models.
\newblock ArXiv:2010.03666.

\bibitem[{D'Elia, {De los Reyes}, and Trujillo(2019)}]{DElia2019imaging}
D'Elia, M.; {De los Reyes}, J.; and Trujillo, A.~M. 2019.
\newblock Bilevel parameter optimization for nonlocal image denoising models.
\newblock ArXiv1912.02347.

\bibitem[{D'Elia et~al.(2017)D'Elia, Du, Gunzburger, and Lehoucq}]{DElia2017}
D'Elia, M.; Du, Q.; Gunzburger, M.; and Lehoucq, R. 2017.
\newblock Nonlocal convection-diffusion problems on bounded domains and
  finite-range jump processes.
\newblock \emph{Computational Methods in Applied Mathematics} 29: 71--103.

\bibitem[{Du, Tao, and Tian(2018)}]{Du2018peridynamic}
Du, Q.; Tao, Y.; and Tian, X. 2018.
\newblock A peridynamic model of fracture mechanics with bond-breaking.
\newblock \emph{Journal of Elasticity} 132(2): 197--218.

\bibitem[{D’Elia and Gunzburger(2016)}]{d2016identification}
D’Elia, M.; and Gunzburger, M. 2016.
\newblock Identification of the diffusion parameter in nonlocal steady
  diffusion problems.
\newblock \emph{Applied Mathematics \& Optimization} 73(2): 227--249.

\bibitem[{Gilboa and Osher(2007)}]{Gilboa2007}
Gilboa, G.; and Osher, S. 2007.
\newblock Nonlocal linear image regularization and supervised segmentation.
\newblock \emph{Multiscale Model. Simul.} 6: 595--630.

\bibitem[{Ha and Bobaru(2011)}]{ha2011characteristics}
Ha, Y.~D.; and Bobaru, F. 2011.
\newblock Characteristics of dynamic brittle fracture captured with
  peridynamics.
\newblock \emph{Engineering Fracture Mechanics} 78(6): 1156--1168.

\bibitem[{Meerschaert and Sikorskii(2012)}]{Meerschaert2012}
Meerschaert, M.; and Sikorskii, A. 2012.
\newblock \emph{Stochastic models for fractional calculus}.
\newblock Studies in mathematics, Gruyter.

\bibitem[{Pang et~al.(2020)Pang, D'Elia, Parks, and
  Karniadakis}]{pang2020npinns}
Pang, G.; D'Elia, M.; Parks, M.; and Karniadakis, G.~E. 2020.
\newblock n{PINN}s: nonlocal {P}hysics-{I}nformed {N}eural {N}etworks for a
  parametrized nonlocal universal {L}aplacian operator. {A}lgorithms and
  {A}pplications.
\newblock To appear in Journal of Computational Physics.

\bibitem[{Pang, Lu, and Karniadakis(2019)}]{Pang2019fPINNs}
Pang, G.; Lu, L.; and Karniadakis, G.~E. 2019.
\newblock f{PINN}s: Fractional Physics-Informed Neural Networks.
\newblock \emph{{SIAM} {J}ournal on {S}cientific {C}omputing} 41: A2603--A2626.

\bibitem[{Scalas, Gorenflo, and Mainardi(2000)}]{Scalas2000}
Scalas, E.; Gorenflo, R.; and Mainardi, F. 2000.
\newblock Fractional calculus and continuous time finance.
\newblock \emph{Physica A} 284: 376--384.

\bibitem[{Schumer et~al.(2003)Schumer, Benson, Meerschaert, and
  Baeumer}]{Schumer2003}
Schumer, R.; Benson, D.; Meerschaert, M.; and Baeumer, B. 2003.
\newblock Multiscaling fractional advection-dispersion equations and their
  solutions.
\newblock \emph{Water Resources Research} 39(1): 1022--1032.

\bibitem[{Silling(2020)}]{Silling2020Pulse}
Silling, S. 2020.
\newblock Propagation of a Stress Pulse in a Heterogeneous Elastic Bar.
\newblock {S}andia Report SAND2020-8197, Sandia National Laboratories.

\bibitem[{Silling(2000)}]{silling2000reformulation}
Silling, S.~A. 2000.
\newblock Reformulation of elasticity theory for discontinuities and long-range
  forces.
\newblock \emph{Journal of the Mechanics and Physics of Solids} 48(1):
  175--209.

\bibitem[{Trask et~al.(2019)Trask, You, Yu, and
  Parks}]{trask2019asymptotically}
Trask, N.; You, H.; Yu, Y.; and Parks, M.~L. 2019.
\newblock An asymptotically compatible meshfree quadrature rule for nonlocal
  problems with applications to peridynamics.
\newblock \emph{Computer Methods in Applied Mechanics and Engineering} 343:
  151--165.

\bibitem[{Weckner and Silling(2011)}]{weckner2011determination}
Weckner, O.; and Silling, S.~A. 2011.
\newblock Determination of nonlocal constitutive equations from phonon
  dispersion relations.
\newblock \emph{International Journal for Multiscale Computational Engineering}
  9(6).

\bibitem[{Xu, D'Elia, and Foster(2020)}]{Xu2020learning}
Xu, X.; D'Elia, M.; and Foster, J. 2020.
\newblock Bond-Based Peridynamic Kernel Learning with Energy Constraint.
\newblock In preparation.

\bibitem[{Xu and Foster(2020)}]{Xu2020deriving}
Xu, X.; and Foster, J. 2020.
\newblock Deriving peridynamic influence functions for one-dimensional elastic
  materials with periodic microstructure.
\newblock ArXiv:2003.05520.

\bibitem[{You, Yu, and Kamensky(2020)}]{you2020asymptotically}
You, H.; Yu, Y.; and Kamensky, D. 2020.
\newblock An asymptotically compatible formulation for local-to-nonlocal
  coupling problems without overlapping regions.
\newblock \emph{Computer Methods in Applied Mechanics and Engineering} 366:
  113038.

\bibitem[{You et~al.(2020)You, Yu, Trask, Gulian, and
  D'Elia}]{You2020Regression}
You, H.; Yu, Y.; Trask, N.; Gulian, M.; and D'Elia, M. 2020.
\newblock Data-driven learning of robust nonlocal physics from high-fidelity
  synthetic data.
\newblock ArXiv:2005.10076.

\end{thebibliography}

\end{document}